\documentclass[conference]{IEEEtran}
\IEEEoverridecommandlockouts
\usepackage{cite}
\usepackage{url} 
\usepackage{amsmath,amssymb,amsfonts}
\usepackage{algorithm}

\usepackage{graphicx}
\usepackage{textcomp}
\usepackage{xcolor}
\usepackage{amsmath}
\usepackage{adjustbox}
\usepackage{booktabs}
\usepackage{algpseudocode}

\def\BibTeX{{\rm B\kern-.05em{\sc i\kern-.025em b}\kern-.08em
    T\kern-.1667em\lower.7ex\hbox{E}\kern-.125emX}}
\begin{document}

\title{Multi-Agent LLM Judge: automatic personalized LLM judge design for evaluating natural language generation applications}

\author{\IEEEauthorblockN{1\textsuperscript{st} Hongliu CAO}
\IEEEauthorblockA{\textit{Amadeus SAS} \\
caohongliu@gmail.com}
\and
\IEEEauthorblockN{2\textsuperscript{nd} Ilias DRIOUICH}
\IEEEauthorblockA{\textit{Amadeus SAS} \\
ilias.driouich@amadeus.com}
\and
\IEEEauthorblockN{3\textsuperscript{rd} Robin SINGH}

\and
\IEEEauthorblockN{4\textsuperscript{th} Eoin  Thomas}
\IEEEauthorblockA{\textit{Amadeus SAS} \\
eoin.thomas@amadeus.com}
}

\maketitle

\begin{abstract}
Large Language Models (LLMs) have demonstrated impressive performance across diverse domains, yet they still encounter challenges such as insufficient domain-specific knowledge, biases, and hallucinations. This underscores the need for robust evaluation methodologies to accurately assess LLM-based applications. Traditional evaluation methods, which rely on word overlap or text embeddings, are inadequate for capturing the nuanced semantic information necessary to evaluate dynamic, open-ended text generation.
Recent research has explored leveraging LLMs to mimic human reasoning and decision-making processes for evaluation purposes known as LLM-as-a-judge framework. However, these existing frameworks have two significant limitations. First, they lack the flexibility to adapt to different text styles, including various answer and ground truth styles, thereby reducing their generalization performance. Second, the evaluation scores produced by these frameworks are often skewed and hard to interpret, showing a low correlation with human judgment.
To address these challenges, we propose a novel dynamic multi-agent system that automatically designs personalized LLM judges for various natural language generation applications. This system iteratively refines evaluation prompts and balances the trade-off between the adaptive requirements of downstream tasks and the alignment with human perception. Our experimental results show that the proposed multi-agent LLM Judge framework not only enhances evaluation accuracy compared to existing methods but also produces evaluation scores that better align with human perception.

\end{abstract}


\section{Introduction}
Large Language Models (LLMs) have demonstrated remarkable performance, leading to their widespread use in various industries \cite{thakur2024judging, cao2024recent, gu2024survey}.
Despite their impressive performance, LLMs face critical challenges such as the absence of domain-specific and updated knowledge, and the prevalence of bias and hallucinations \cite{cao2024writing, gallegos2024biassurvey, wu2023style}. These challenges highlight the necessity for robust evaluation methodologies to assess the performance of LLM-based applications. 
However, the automatic evaluation of text generation quality across diverse tasks, especially for free-form text responses, remains a significant challenge \cite{xu2023instructscore, thakur2024judging, chang2024survey}. 
Classic n-gram matching-based evaluation methods, such as BLEU \cite{papineni2002bleu} and ROUGE \cite{lin2004rouge}, which measure word overlap between generated and reference texts, are widely employed but prove inadequate for dynamic, open-ended scenarios \cite{reiter2018structured,li2024generation}.
The development of semantic text embedding models, including Bidirectional Encoder Representations from Transformers (BERT) \cite{devlin2018bert}, has facilitated the development of embedding-based metrics like BERTScore \cite{zhang2019bertscore}. Although these metrics have improved semantic understanding, they still face challenges in accurately capturing nuanced semantic information \cite{hossain2022analysis}.

The extensive use of generative models like GPT \cite{achiam2023gpt} has further highlighted the capabilities of LLMs in various aspects, including natural language understanding, instruction-following, and in-context learning \cite{cao2024recent, cao2024writing}. The advancement in this field has sparked a surge of interest among scholars to leverage the capabilities of LLMs to emulate human-like reasoning and decision-making processes for the purpose of evaluating the responses/answers generated by applications based on LLMs, a concept commonly referred to as "LLM-as-a-judge".
Traditional evaluation methods require significant expert effort and are often hindered by time constraints and the high costs of qualified evaluators \cite{gu2024survey}. The LLM-as-a-judge framework offers a cost-effective and scalable alternative to human evaluators  by automating the evaluation process \cite{li2024generation}.

\begin{figure}[h]
  \centering
  \includegraphics[width=0.88\linewidth]{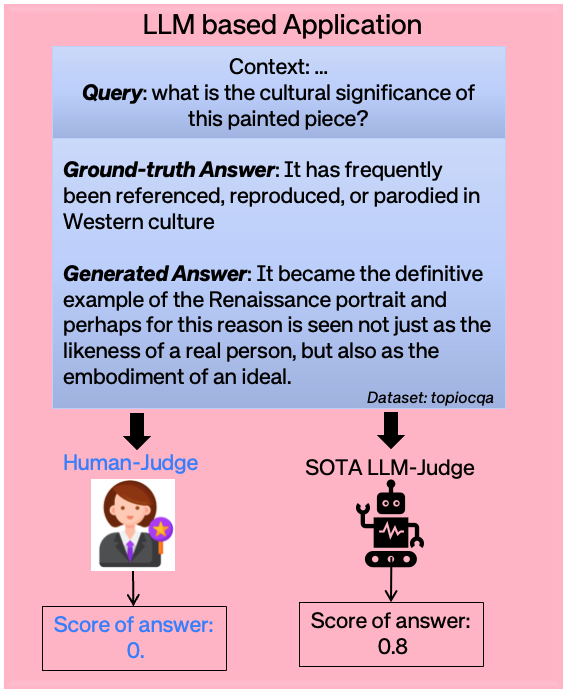}
  \caption{Comparison of answer correctness between human judge and an advanced LLM judge for a given query, ground-truth answer from TopicQA \cite{adlakha2022topiocqa}, and LLM generated answer. While human judges can easily identify the generated answer as incorrect, the state-of-the-art LLM judge fails to recognize this simple error. } 
  \label{fig:eg}
\end{figure}

Several LLM-as-a-Judge frameworks, such as RAGAS \cite{es2023ragas} and Continuous-Eval (CE) \cite{ce}, have gained widespread adoption for the evaluation of Question Answering (QA) systems. 
These frameworks employ a critique LLM to assess the responses generated from various applications. 
In these frameworks, the evaluator LLM is provided with the generated answer and the ground truth answer, and in some cases, the question as well. The evaluator LLM then uses well-defined scoring rubrics and well-engineered prompts to score/grade the responses. 
However, these existing frameworks exhibit two main limitations. Firstly, they lack the flexibility to adapt to different text styles including answer styles or ground truth styles from various natural language generation applications, leading to reduced generalization performance. Secondly, the evaluation scores generated by these frameworks are often skewed and often difficult to understand, with a low correlation to human perception as shown in the example of Figure \ref{fig:eg}.

To address these limitations, we propose a novel dynamic, multi-agent system to design personalized LLM judges for different natural language generation applications automatically. Initially, to align the evaluator score with human perception, we provide the definition and rubrics of human-perceived semantic similarity in the original prompt. Subsequently, the Sample Selection agent selects a small, diverse set of examples for the Evaluation agent. The Evaluation agent then uses the original prompt and selected examples as input, providing feedback to enhance the original prompt. Finally, the ReWrite agent uses the original prompt and feedback from the Evaluation agent to propose improved prompts for the LLM judge. This iterative improvement process continues until either the predefined performance is achieved or the maximum number of iterations is reached.
Through this iterative process, the proposed framework 
effectively balances the trade-off between the predefined semantic similarity criteria and the adaptive requirements of downstream tasks, resulting in a more robust and contextually appropriate alignment with human perception.
Our experimental results demonstrate that the proposed multi-agent LLM Judge framework not only improves evaluation accuracy compared to existing solutions but also offers evaluation scores that align better with human perception.

\section{Related works}

LLM-as-a-Judge frameworks have been used to evaluate various Natural Language Processing (NLP) and Natural Language Understanding (NLU) tasks including Retrieval-Augmented Generation (RAG) \cite{huang2024survey}, code comprehension \cite{yuan2023evaluating}, machine translation \cite{bavaresco2024llms} and more general open-ended tasks \cite{zheng2023judging}. 
These frameworks are designed to assess a range of specific attributes, including but not limited to, correctness, faithfulness, helpfulness, harmlessness, reliability, relevance, feasibility, and overall quality \cite{adlakha2023evaluatinginstqa , li2024generation}. 
The output of LLM-as-a-Judge frameworks can take various forms, such as a continuous or discrete score, a ranking of potential answers, or solutions to true/false or multiple-choice questions \cite{gu2024survey, li2024generation}.

The performance of LLM-as-a-Judge frameworks is often undermined by the variability in ground truth answer styles across different evaluation tasks and the diverse answer styles inherent to various applications based on different LLM models \cite{cao2024writing}. Additionally, inherent biases in LLMs, such as length, positional, and concreteness biases, further compromise evaluation results \cite{park2024offsetbias}. Consequently, enhancing the performance of LLM-as-a-Judge frameworks remains a significant challenge for their effective use as evaluators. To tackle these issues,  state-of-the-art solutions can be broadly categorized into two groups: tuning approaches and prompting approaches \cite{li2024generation}.

\textbf{Tuning approaches:}
The LLM-as-a-judge framework requires that judge LLMs possess the evaluative capacities to comprehend natural language, learn from in-context examples, follow instructions consistently, reason effectively, and align with human perception. However, even advanced LLMs such as GPT4 encounter challenges like conceptual confusion \cite{hu2024llm}. This issue is even more pronounced in smaller open-source LLMs which are significantly limited in their evaluation capabilities despite being easier to implement as evaluators \cite{gu2024survey}. Consequently, many state-of-the-art studies suggest fine-tuning these LLMs to enhance their evaluative capacities.

Supervised fine-tuning (SFT) is a widely used method to enhance the evaluation abilities of judge LLMs \cite{li2024generative, xie2024sorry,yue-etal-2023-automatic}. 
For instance, INSTRUCTSCORE \cite{xu2023instructscore} aims to produce high-quality scores and detailed diagnostic reports for candidate texts by iteratively fine-tuning the 7B LLaMA model \cite{touvron2023llama7b} using both explicit human instructions and automatic feedback from GPT-4 on identified failure modes. 
Vu et al. have developed the Foundational Large Autorater Models (FLAMe), which are trained through supervised multitask fine-tuning on 102 quality assessment tasks, incorporating over 5.3 million human judgments standardized from publicly available evaluations in previous research \cite{vu2024foundational}. 
To enable LLM judges to generalize across various evaluation aspects, Liu et al. \cite{liu-etal-2024-x} propose a two-stage instruction tuning framework called X-EVAL. The first stage involves vanilla instruction tuning to improve the judge model's instruction-following ability, while the second stage focuses on advanced instruction tuning to exploit the
connections between fine-grained evaluation aspects  \cite{liu-etal-2024-x}. For improving the quality of hallucination judges, Wang et al. \cite{wang2024halu} propose to use both supervised fine-tuning and fine-tuning with Directed Preference Optimization (DPO) \cite{rafailov2024direct} in a multiple-evidence setting. 

\textbf{Prompting approaches:}

An evaluation prompt serves as a crucial input for LLM-as-a-judge frameworks, guiding them to execute specific evaluation tasks. 
LLMs exhibit instruction following and in-context learning capabilities, enabling them to perform designated tasks by interpreting examples or instructions embedded within prompts, without necessitating weight updates or retraining \cite{brown2020language}. More importantly, prompting strategies can serve as effective tools in mitigating inherent biases \cite{zheng2023judging}. This underscores the pivotal role of evaluation prompt design in  enhancing the performance of LLM-as-a-judge \cite{gu2024survey}. 

By analyzing generic quality prompt, criteria specific prompt and full rubric prompt with increasing levels of instructions about the target quality of an evaluation,  the authors in in \cite{murugadoss2024evaluating} conclude that full rubric information helps for non-default textual quality evaluations.
\cite{liu2023calibrating} proposes capturing human preferences through human-provided labels, querying LLMs to draft initial scoring criteria via in-context learning, and refining the best-performing criteria through self-improvement. \cite{dong2024can} examines the reliability of LLM-as-a-personalized-judge, which incorporates persona-based principles, and suggests enhancing this framework with verbal uncertainty estimation.
For few-shot example selection, JADE \cite{zhang2023jade} employs human judges to correct LLM evaluations and updates the example sets with the most frequently corrected samples. 
In \cite{li2023mot} the authors propose to prompt LLMs to retrieve appropriate demonstrations based on the candidates’ relevance in solving specific problems. 
There are also several recent studies focusing on enhancing the performance of LLM judges through multi-LLMs collaboration approaches. Li et al. \cite{li2023prd} introduce two notable methods: the Peer Rank (PR) algorithm, which considers each peer LLM's pairwise preferences to generate a final ranking of models, and Peer Discussion (PD), where two LLMs engage in a dialogue to reach a consensus on answer preferences. Additionally, Jung et al. \cite{jung2024trust} propose the Cascaded Selective Evaluation method, which begins with a smaller, cost-effective model to make initial judgments, assesses its confidence, and escalates to a stronger model only when necessary.

Due to the fast adoption of LLM based applications such as Retrieval Augmented Generation (RAG) systems, several LLM-as-a-judge frameworks have been proposed to contribute to faster evaluation cycles of RAG architectures. Among these, the most adopted are RAGAS \cite{es2023ragas} and Continuous-Eval (CE) \cite{ce}. RAGAS provides various evaluation metrics including faithfulness, answer relevancy, answer correctness, etc. The answer correctness from RAGAS takes a weighted average of the semantic similarity and the argument-based factual similarity measured by LLMs to arrive at the final score.  
Continuous-Eval instead measures the answer correctness with a single score leveraging few-shot examples and detailed evaluation rubrics for each of the scores.

In summary, LLM-as-a-Judge frameworks are increasingly being utilized to evaluate a wide array of NLP and NLU tasks, their effectiveness is often hindered by diversity in text styles and inherent biases within LLMs. To address these issues, current research has focused on two primary approaches: tuning and prompting. Tuning methods, such as supervised fine-tuning and advanced instruction tuning, aim to enhance the evaluative capacities of LLMs by refining their performance on specific tasks. On the other hand, prompting strategies leverage the instruction-following and in-context learning capabilities of LLMs to guide them in executing evaluation tasks more effectively, which is more sustainable and cost-efficient than tuning methods. 
Despite these advancements, existing frameworks still face  two major limitations as illustrated in the example in Figure \ref{fig:eg}. First, they lack the flexibility to adapt to different text styles, including answer styles (e.g. different LLMs have different answer styles) or ground truth styles (e.g. some applications have single ground truth answer while others can provide multiple ground truth answers that are all correct) from various natural language generation applications, resulting in poorer generalization performance. Second, the evaluation scores produced by these frameworks are often skewed and difficult to interpret, showing a low correlation with human judgment, which hinders the meaningful interpretation of these evaluation scores.

\section{The proposed solution}
\begin{figure}[h]
  \centering
  \includegraphics[width=0.96\linewidth]{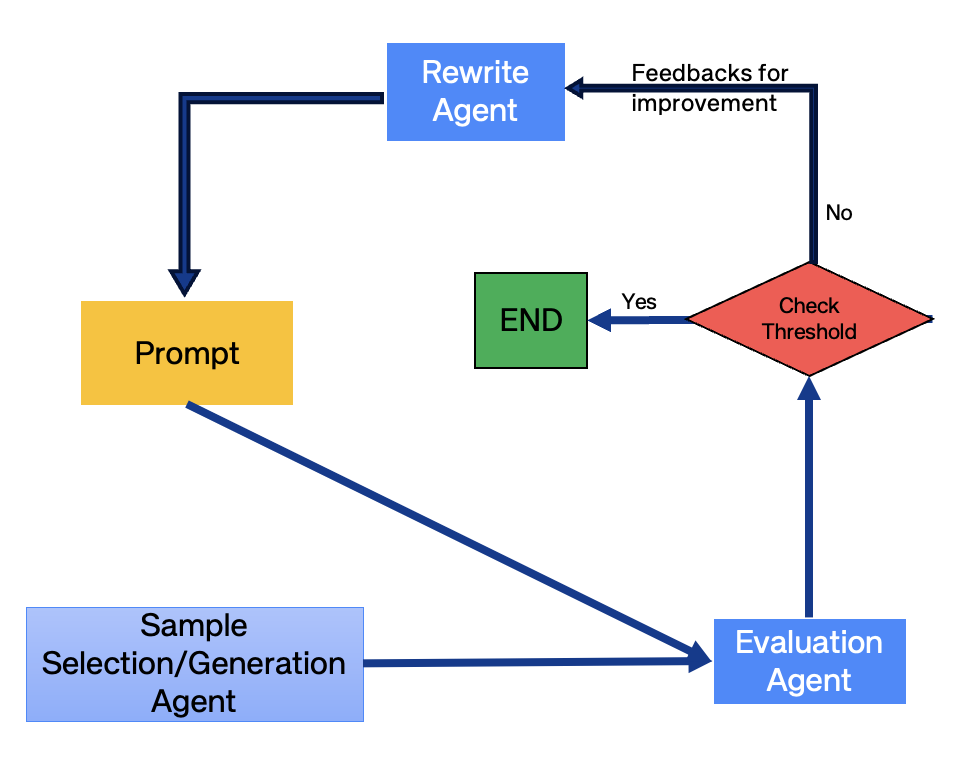}
  \caption{The proposed multi-agent LLM judge framework operates through the following workflow: Initially, the Prompt block contains the Initial Prompt, which can be updated in later phases. The Sample Selection agent's role is to select a diverse and representative set of examples for the Evaluation agent. The Evaluation agent tests these examples against the input prompt, providing an overall evaluation score as well as detailed feedback for improving the input prompt. The ReWrite agent then reviews both the input prompt and the feedback from the Evaluation agent to produce revised prompts that better guide the LLM judge. The iteration loop continues until the evaluation score meets the user's requirements or the maximum number of iterations is reached. } 
  \label{fig:flow}
\end{figure}

In this study, we introduce a novel dynamic multi-agent system designed to automatically create personalized LLM judges for various natural language generation tasks, without the need for crafting large datasets or extensive tuning of the LLMs. 
The overall workflow of our proposed solution is illustrated in Figure \ref{fig:flow}. 
To ensure that the LLM judge's scores are aligned with human perception and easy to understand, we incorporate definitions of human-perceived semantic similarity into the initial prompt. 
The evaluation rubrics from established Semantic Textual Similarity (STS) literature such as \cite{agirre2013sem, stsbcer2017semeval, cao2024writing} are used in the Initial Prompt:
 \begin{itemize}
     \item 0 means the the pair of texts are on different topics;
     \item 0.2 means the the pair of texts are not equivalent, but are on the same topic;
     \item 0.4 means  the the pair of texts are not equivalent, but share some details;
     \item 0.6 means  the the pair of texts are roughly equivalent, but some important information differs/missing;
     \item 0.8 means  the the pair of texts are mostly equivalent, but some unimportant details differ; 
     \item 1 means  the the pair of texts are completely equivalent;
 \end{itemize}

Subsequently, the Sample Selection agent is responsible for curating a diverse and representative set of examples for the Evaluation agent. If there is a training dataset which contains queries, ground-truth answers and generated answers related to a specific downstream task,  text clustering techniques are employed to organize the dataset into distinct clusters. A single example from each cluster is then selected randomly to compose the final few-shot examples for the Evaluation agent. The primary objectives of the Sample Selection agent are twofold: (1) to choose representative and diverse examples that enable the LLM judge to adapt to various text styles, and (2) to minimize redundancy and token size of the few-shot examples, considering the context length limitations, speed, cost, and sustainability. 
In scenarios where ground-truth question-answer pairs are unavailable, such as in new industrial Question-Answering applications or Retrieval-Augmented Generation (RAG) systems, small-sized few-shot examples can be generated either by humans or by LLMs.

The Evaluation agent takes the prompt and selected examples as input. Unlike previous studies that incorporate the selected few-shot examples directly into the prompt, our proposed Evaluation agent tests these examples against the prompt. Then it provides detailed feedback specifically on the mistaken examples to refine and enhance the input prompt. 

Finally, the ReWrite agent carefully examines the input prompt along with the feedback provided by the Evaluation agent. By analyzing this information, the ReWrite agent identifies areas where the prompt can be improved. It then automatically generates revised and more effective prompts that are specifically designed to better guide the LLM judge.

With the help of the Evaluation agent and the ReWrite agent, the proposed multi-agent LLM judge framework iteratively assesses its prompt (initial or updated) against the selected or generated few-shot examples, systematically incorporating complementary information from these examples and feedback into its prompt. Through this iterative process, the proposed framework effectively balances the trade-offs between predefined semantic similarity criteria and the adaptive needs of downstream tasks. This leads to a more robust and contextually appropriate alignment with human perception.
Based on the evaluation results, the Evaluation agent also provides an overall score for the original prompt. This score is then compared against a predefined threshold: if the score falls short of the user's requirements, the feedback is forwarded to the ReWrite  agent. If the score meets the user's requirements or the maximum number of iterations is reached, the iteration loop terminates as shown in Algorithm \ref{alg:llm-judge}.

\begin{algorithm}
\caption{Proposed Multi-Agent LLM judge framework}
\label{alg:llm-judge}
\begin{algorithmic}[1]
\Require Initial prompt $P_0$, evaluation threshold $T$, maximum iterations $I_{\max}$
\\
\Return Final Prompt $P_{\text{final}}$

\State $P \gets P_0$ \Comment{Initialize the prompt}
\Repeat
    \State \textbf{Sample Selection Agent:}
    \State Partition dataset $\mathcal{D}$ into clusters $\{\mathcal{C}_j\}_{j=1}^C$
    \State Select one or more representative example $e_j \sim \mathcal{C}_j$ for each cluster
    \State $E \gets \{e_j\}_{j=1}^C$  

   \State \textbf{Evaluation Agent:}
    \State 1. Evaluate the selected examples E against the prompt P and get the average evaluation score: $S(P, E)$ 
    \State 2. Generate feedback from the example evaluations $F = \text{GenerateFeedback}(P, E, S(P, E))$

    \If{$S(P, E) \geq T$}
        \State Terminate interation
    \EndIf

    \State \textbf{ReWrite Agent: Update the prompt}
    \[
    P \gets \text{ReWrite}(P, F)
    \]

\Until{ Maximum iteration $I_{\max}$ is reached}
\State \textbf{Return:} $P_{\text{final}} \gets P$
\end{algorithmic}
\end{algorithm}

\section{Experiments}
\subsection{Research Question 1: How accurate is the proposed multi-agent LLM judge?}

The proposed multi-agent LLM judge framework aims to achieve two primary goals: (1) enhancing the performance and adaptability of the LLM-as-a-judge system to different text styles from different natural language generation tasks, and (2) improving the alignment between the evaluation scores generated by the LLM judge and those provided by human annotators. In this section, we will assess the proposed solution with respect to the first objective. The evaluation of the second objective will be covered in the subsequent section.

\subsubsection{Dataset}
In this section, we utilize the Instruct-QA dataset \cite{adlakha2024evaluating} due to its inclusion of a variety of task types, distinct ground-truth styles, and diverse generated answer styles from various LLMs. The Instruct-QA dataset encompasses three different information-seeking Question-Answering tasks (the overall statistics of these three datasets are shown in Table \ref{sts}.), including:
\begin{itemize}
    \item Open-domain QA task: Natural Questions dataset \cite{lee2019latent} with queries from Google search engine.
    \item Multi-hop QA task:  HotpotQA dataset \cite{yang2018hotpotqa} with  at least two
Wikipedia passages to reason upon jointly. 
    \item Conversational QA task: TopiOCQA dataset \cite{adlakha2022topiocqa} with open-domain information-seeking dialogue. 
\end{itemize}

\begin{table}[h!]
\caption{ The datasets' statistics of Instruct-QA.  Answer length is the average number of words. The validation splits are used in Instruct-QA  \cite{adlakha2023evaluatinginstqa } as the test sets are hidden.}\label{sts}
\begin{adjustbox}{width=0.48\textwidth}
\centering
\begin{tabular}{l rrr}
\toprule
Dataset & \#Questions & Answer length  & \#Passages (millions)  \\
\midrule
Natural Questions & 3,610 & 2.16 & 21\\ \hline
HotpotQA & 7,405 & 2.46 &5.2\\ \hline
TopiOCQA & 2,514 & 10.98 &25.7\\ \hline
\end{tabular}
\end{adjustbox}
\end{table}

 RAGs based on a standardized prompt template are used to generate the answers for the queries with 4 different LLMs including 
 FlanT5-xxl \cite{chung2024scalingt5} with 11B parameters, Alpaca-7b \cite{taori2023stanfordalpaca}, GPT3.5-turbo and Llama2-7b \cite{touvron2023llama}. 
 The correctness of each generated answer is annotated by human evaluators. 

\subsubsection{Experimental protocol}
In this study, we select two widely used LLM-as-a-judge frameworks  as our baselines: RAGAS \cite{es2023ragas} and Continuous-Eval (CE) \cite{ce}. The RAGAS framework determines the correctness of answers by calculating a weighted average of two key factors: the semantic similarity between the generated answer and the reference answer and the factual similarity based on arguments measured by LLM judge to arrive at the final score. To have a fair comparison of LLM-as-a-judge frameworks, only LLM judge part is used. CE measures the answer correctness leveraging few-shot examples and detailed evaluation rubrics. 

GPT-3.5 Turbo has been chosen as the foundational model for all LLM judges because it is widely used and has demonstrated reliable performance. Additionally, the more advanced GPT-4 model is employed by the ReWrite agent to create improved prompts based on feedback, enhancing the overall quality of the LLM judge. 
To ensure consistency and reduce randomness in the outputs, the temperature parameter for all LLM judges is set to 0. 
The Instruct-QA data contain three diverse datasets from which the Sample Selection agent randomly selects one example per dataset. The maximum iteration number is set to 10 for the proposed multi-agent LLM judge. 
The area under Receiver Operating Characteristic (ROC) curve is used as the evaluation metric to measure the performance of different LLM judges.

\subsubsection{Experimental results}
To evaluate how effective the proposed multi-agent LLM judge is as well as the utility of different agents, we conduct a thorough analysis of its performance. This analysis include two specific conditions for comparison. First, we measure the performance of the initial prompt before any iterative improvements are made. This condition is referred to as the "Initial Prompt".  Second, we assess the performance of the initial prompt when it is simply combined with the few-shot examples selected by the Sample Selection Agent. This condition is referred to as the "Few-shot Prompt".  By comparing these two conditions, we aimed to understand how much the iterative improvements contribute to the overall effectiveness of the proposed multi-agent LLM judge.

\begin{figure}[h]
  \centering
  \includegraphics[width=0.98\linewidth]{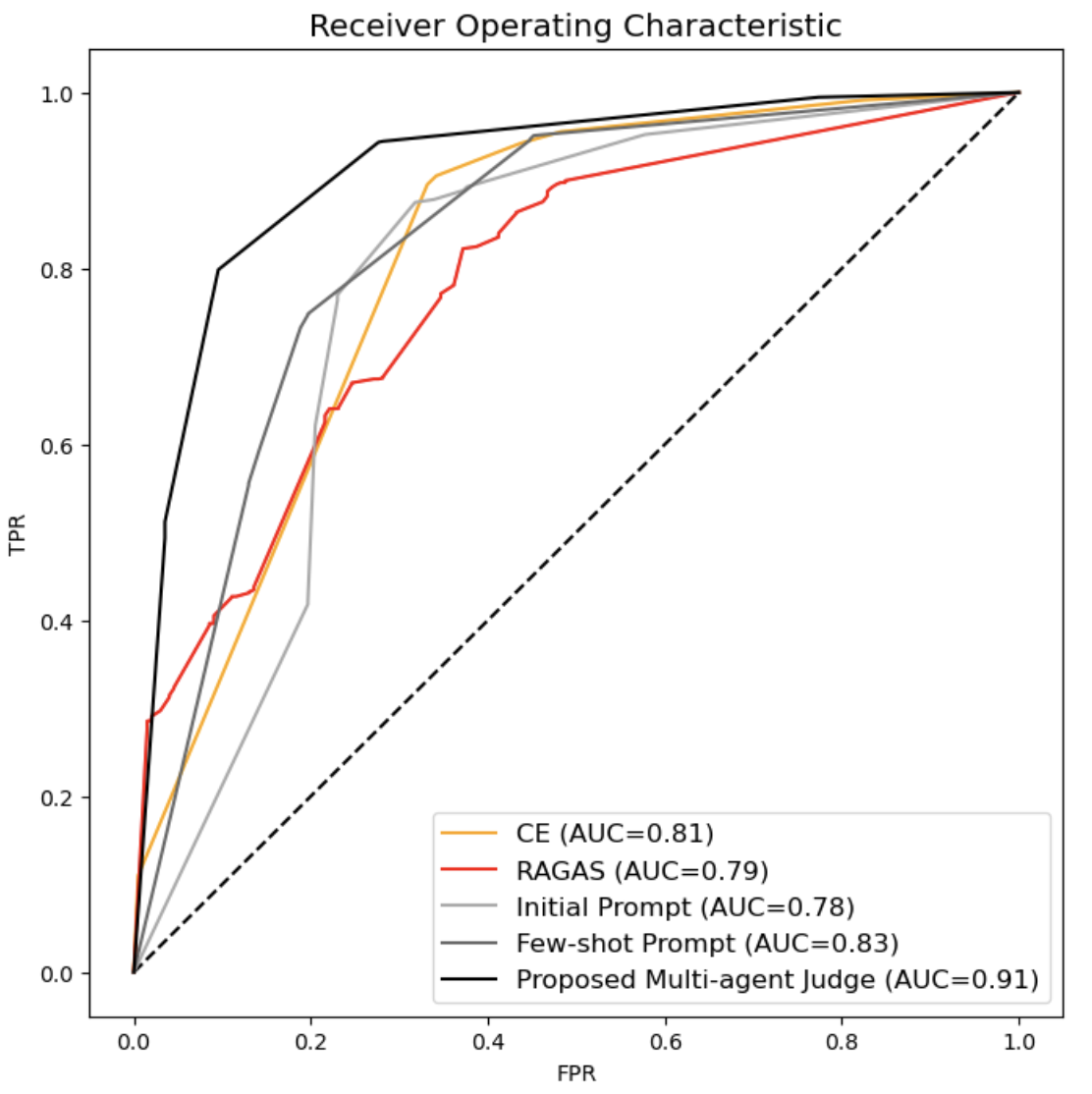}
  \caption{The experimental results on Instruct-QA  datasets of different LLM judges: the X-axis denotes the False Positive Rate (FPR), and the Y-axis indicates the True Positive Rate (TPR). Each method's ROC curve is depicted in a distinct color, with the corresponding Area Under the Curve (AUC) values displayed in the bottom right corner of the figure. } 
  \label{fig:res1}
\end{figure}

The experimental results on Instruct-QA  datasets of different LLM judges are shown in Figure \ref{fig:res1}: the X-axis denotes the False Positive Rate (FPR), and the Y-axis indicates the True Positive Rate (TPR). Each method's ROC curve is depicted in a distinct color, with the corresponding Area Under the Curve (AUC) values displayed in the bottom right corner of the figure. 
It can be observed that the Initial Prompt method has the lowest performance with an AUC value of 0.78. This outcome is expected, as the Initial Prompt includes only basic information, such as the task description and semantic similarity rubrics without any prompt engineering.
Notably, the widely adopted RAGAS framework showed only a marginal improvement, achieving an AUC of 0.79. In contrast, the Continuous-Eval (CE) method has better performance than both the Initial Prompt and RAGAS frameworks with an AUC value of 0.81.

The Few-shot Prompt method, which simply combines the Initial Prompt with few-shot examples chosen by the Sample Selection agent, surpasses the performance of the well engineered RAGAS and CE frameworks with an AUC value of 0.83. This outcome highlights the limitations of the state-of-the-art LLM-as-a-judge frameworks in adapting to diverse text styles across various natural language generation applications, resulting in diminished generalization performance. Furthermore, these results underscore the efficacy of simply incorporating few-shot examples.

To better integrate the Initial Prompt consisting of human defined rubrics of semantic similarity (in order to align better with human perception) with the downstream few-shot examples (in order to adapt better to different downstream tasks and text styles), the proposed multi-agent LLM judge makes improvements iteratively instead of simply adding few shot examples to the prompt. The multi-agent LLM judge iteratively assesses the its prompt (initial or updated) against the few-shot examples, systematically incorporating complementary information from these examples and feedback into its prompt. Through this iterative process, the proposed framework effectively balances the trade-off between the predefined semantic similarity criteria and the adaptive requirements of downstream tasks, resulting in a more robust and contextually appropriate alignment with human perception. As illustrated in Figure \ref{fig:res1}, the proposed multi-agent LLM judge demonstrates superior performance with an AUC value of 0.91, which is much better than simply merging the initial prompt with the few-shot examples.

\subsection{Research Question 2: How well is the proposed multi-agent LLM judge aligned with human perception?}
In the experiments and analysis described in the previous section, we have assessed the performance of our proposed solution by comparing it to four baseline solutions and showed that the proposed solution outperformed all the baselines.
In this section, we aim to address our second objective: to evaluate whether the proposed multi-agent LLM judge can improve the alignment between the evaluation scores generated by the LLM judge and those provided by human annotators. 

\subsubsection{Dataset}
The Semantic Textual Similarity Benchmark (STSB) \cite{stsbcer2017semeval} is selected in this section as it 
provides a wide range of human annotation scores rather than limiting the annotations to binary or categorical labels. 
STSB is a collection of English datasets, which have been utilized in the *SEM and SemEval STS shared tasks spanning from 2012 to 2017 \cite{stsbcer2017semeval}. The annotation of the similarity between pairs of texts is achieved through a crowdsourcing approach, incorporating both pragmatic and global knowledge. 
The diversity in human annotated scores in STSB allows for a more nuanced comparison between the scores produced by LLM judges and those given by human evaluators, thereby enabling a more thorough and meaningful evaluation of the alignment between the two sets of scores.

\subsubsection{Experimental protocol}
The same baselines as the previous sections are compared in this section. All the prompts from different LLM judges are also kept the same as in the previous section. However, to  measure if the score generated by different LLM judges correlates well with human annotation, the Pearson correlation is used as the evaluation metric in this section following the evaluation protocol of STSB. 

\subsubsection{Experimental results}

 \begin{figure}[h]
  \centering
  \includegraphics[width=\linewidth]{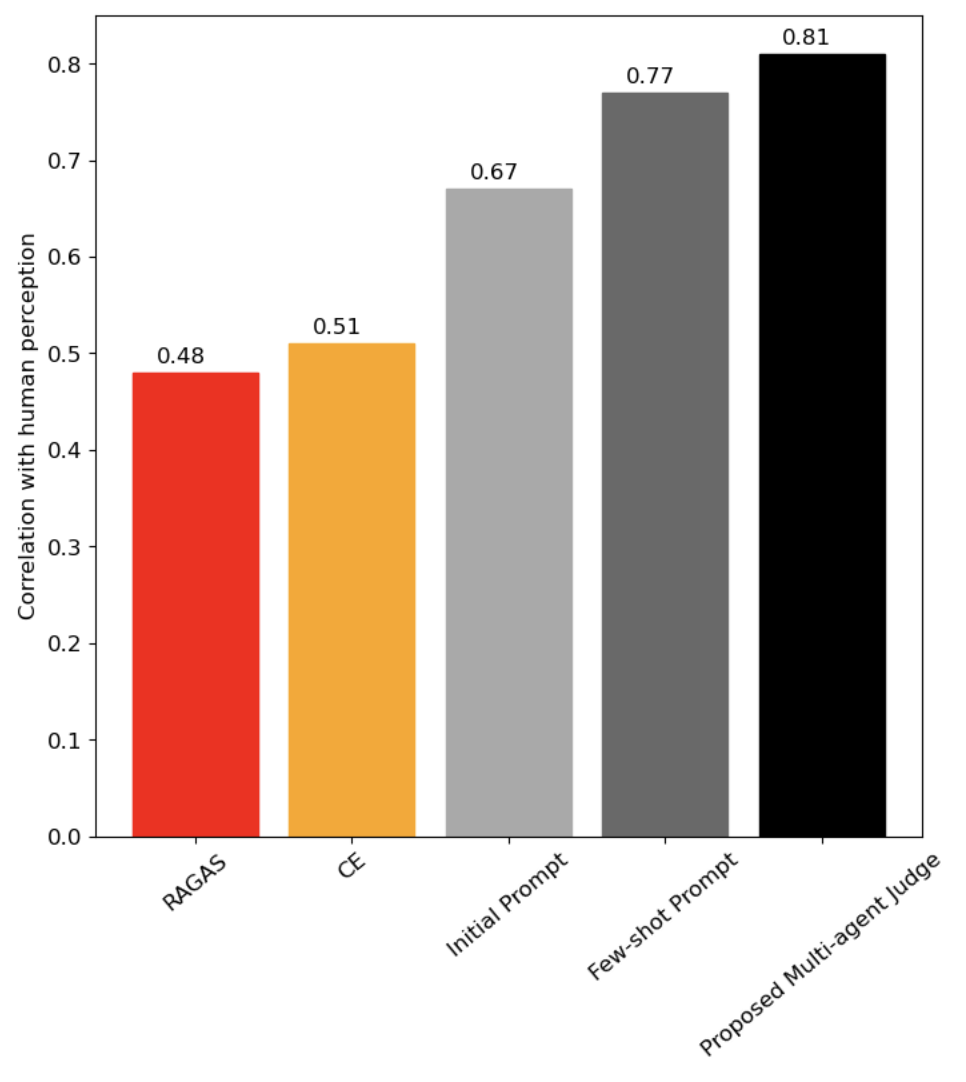}
  \caption{ Evaluation of the alignment between LLM judges and human perception: Pearson correlation between scores generated by LLM judges and human annotations are shown in this figure:  the X-axis denotes different LLM judges and  the Y-axis denotes the correlation score with human annotations.} 
  \label{fig:res2}
\end{figure}

The experimental results on STSB of different LLM judges are shown in Figure \ref{fig:res2}:  the X-axis denotes different LLM judges and the Y-axis denotes the correlation score with human annotations. It is evident from the figure that the RAGAS score has the lowest correlation with human annotations, suggesting the poorest alignment with human perception. The CE score demonstrates a marginally better alignment than the RAGAS score, but it still shows a low correlation with human annotations, with a correlation value of 0.51.

The experimental results discussed in the previous section show that the Initial Prompt has the lowest accuracy on the Instruct-QA dataset, with an AUC value of 0.78 (see Figure \ref{fig:res1}). This is because the Initial Prompt proposed in this study is simply given definitions of human-perceived semantic similarity, aiming to make the generated scores more consistent with human annotations. However, when it comes to the alignment task on the STSB dataset, the Initial Prompt performs much better than both the RAGAS score and CE score, achieving a correlation value of 0.67. 
The Few-shot Prompt further increases the correlation score of the Initial Prompt to 0.77. However, the proposed multi-agent LLM judge achieves an even higher alignment with human annotations, with a correlation score of 0.81. These findings indicate that the proposed multi-agent LLM Judge framework not only enhances the accuracy of LLM judges but also improves the alignment between the evaluation scores generated by the LLM judge and those provided by human annotators.

\subsection{Research Question 3: What does the proposed multi-agent LLM judge improve upon the Initial Prompt? }

\begin{figure}[h]
  \centering
  \includegraphics[width=\linewidth]{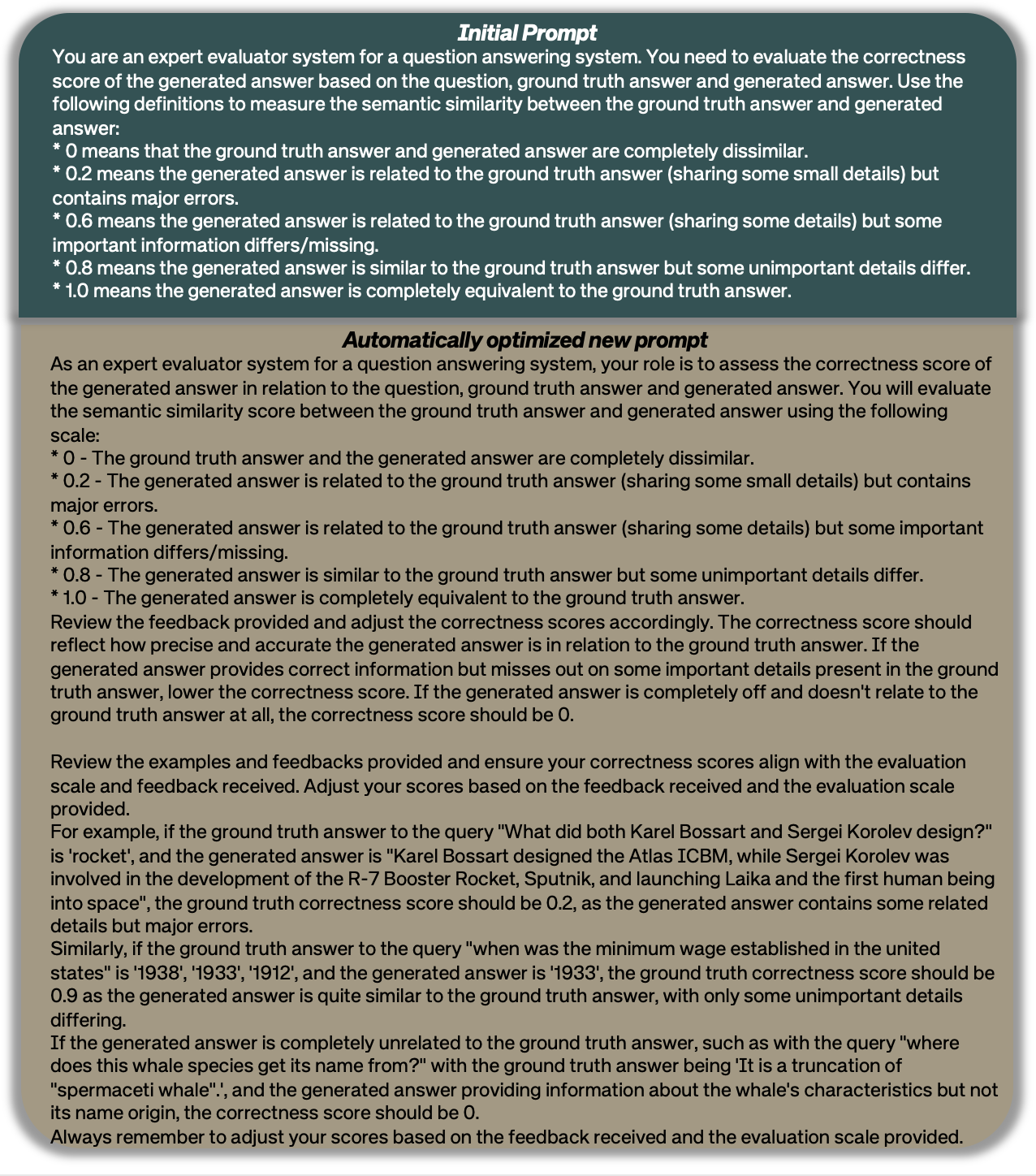}
  \caption{A comparison between the Initial Prompt (displayed at the top) and the automatically optimized final prompt generated by the proposed multi-agent LLM judge (shown at the bottom).} 
  \label{fig:prompt}
\end{figure}

From the experimental results in the previous two sections, it can be concluded that the proposed multi-agent LLM judge can generate more accurate and meaningful evaluation scores that aligns better with human perception. To explore the reasons behind its superior performance, we conduct a detailed comparison between the Initial Prompt and the optimized final prompt generated by the multi-agent LLM judge. The results of this comparison are illustrated in Figure \ref{fig:prompt}.
The optimized final prompt is more extended and detailed. It retains the clear definitions of the semantic similarity scores from the Initial Prompt but adds more context and instructions, with the focus on including examples, additional instructions on adjusting scores based on feedback, and a more detailed explanation of the evaluation process. 

The key enhancements from the automatically optimized final prompt from the proposed multi-agent LLM judge can be summarized as:
\begin{itemize}
    \item The optimized prompt provides more detailed and comprehensive instructions, reducing ambiguity and ensuring that the designed LLM judge understands how to apply the scoring system correctly.
    \item By including specific examples, the optimized prompt helps the designed LLM judge visualize how to apply the scoring system in different scenarios, leading to more accurate and consistent evaluations.
    \item Emphasizing the importance of feedback and providing instructions on adjusting scores based on feedback ensures that the designed LLM judge continuously improves their scoring accuracy, leading to better overall performance.
    \item The additional guidance on how to handle different situations and the emphasis on precision and accuracy help the designed LLM judge make more informed decisions, resulting in more reliable evaluations.
    \item Repeating key concepts and instructions, such as the need to adjust scores based on feedback, reinforces these ideas and ensures that they are consistently applied.
\end{itemize}

\section{Conclusion and future work}
As the number of natural language generation applications continues to rise, it becomes increasingly crucial to establish effective methods to  evaluate the performances of these applications. While human evaluators can be both time-consuming and expensive, the LLM-as-a-judge framework offers a cost-effective and scalable alternative. However, existing LLM judges struggle to generalize to different text styles and often fail to produce scores that accurately reflect human judgment.
To tackle these challenges, a dynamic multi-agent system is introduced in this work to automatically create personalized LLM judges for different natural language generation tasks. The proposed framework continuously improves the evaluation prompt while balancing the adaptation to downstream tasks and alignment with human judgment. Our experiments demonstrate that the proposed multi-agent LLM Judge framework not only improves evaluation accuracy over existing solutions but also generates scores that align better with human perception. By analyzing the difference between the Initial Prompt and the optimized final
prompt generated by the multi-agent LLM judge, we also provide insights into more effective prompt design. However, we have focused exclusively on two main components of the LLM-as-a-judge framework: its accuracy and how well it aligns with human perception. Future research will explore additional factors such as faithfulness, harmlessness, reliability, and biases.

\section{Ethical Considerations}
This study investigates the potential of using multi-agent system to automatically improve the accuracy and human perception alignment of LLM judges. To carry out our research, we utilized LLMs and data sets that are publicly accessible, ensuring our experiments did not raise any ethical concerns. 

\bibliographystyle{ieeetr}
\bibliography{sample-base}

\end{document}